\pdfoutput=1
\documentclass[letterpaper, 10 pt, conference]{ieeeconf}  

\IEEEoverridecommandlockouts                              

\usepackage{amsmath}
\usepackage{todonotes}
\usepackage{array}
\usepackage{breqn}

\definecolor{pinegreen}{rgb}{0.0, 0.47, 0.44}

\newenvironment{conditions}
  {\par\vspace{\abovedisplayskip}\noindent\begin{tabular}{>{$}l<{$} @{${}={}$} l}}
  {\end{tabular}\par\vspace{\belowdisplayskip}}

\title{\LARGE \bf Automated Camera-Based Estimation of Rehabilitation Criteria Following ACL Reconstruction
}

\author{Choong Hee Kim, Shannon M. Danforth, Patrick D. Holmes, Daphna Raz, \\ Darlene Yao, Asheesh Bedi, and Ram Vasudevan
\thanks{This work is supported by the National Science Foundation Career Award \#1751093, Michigan Medicine Fostering Innovation Grant \#U045830, Michigan Integrative Musculoskeletal Health Core Center Grant AR-069620, and Michigan Institute for Clinical \& Health Research Grant UL1TR000433. }
\thanks{C.H. Kim, S.M. Danforth, P.D. Holmes, D. Raz, D. Yao, A. Bedi, and R. Vasudevan are at the University of Michigan,
        Ann Arbor, MI, USA.
        {\tt\small \{frankkim, sdanfort, pdholmes, daphraz\} @umich.edu, \{dyao, abedi\}@med.umich.edu, and \{ramv\}@umich.edu}}%
}

\begin{document}

\maketitle
\thispagestyle{empty}
\pagestyle{empty}

\begin{abstract}

Anterior cruciate ligament (ACL) reconstruction necessitates months of rehabilitation, during which a clinician evaluates whether a patient is ready to return to sports or occupation.
Due to their time- and cost-intensive nature, these screenings to assess progress are unavailable to many.
This paper introduces an automated, markerless, camera-based method for estimating rehabilitation criteria following ACL reconstruction.
To evaluate the performance of this novel technique, data were collected weekly from 12 subjects as they used a leg press over the course of a 12-week rehabilitation period. 
The proposed camera-based method for estimating displacement and force was compared to encoder and force plate measurements.
The leg press displacement and force values were estimated with 89.7\% and 85.3\% accuracy, respectively. These values were then used to calculate lower-limb symmetry and to track patient progress over time.

\end{abstract}

\section{Introduction}

Rupture of the anterior cruciate ligament (ACL) is a common lower limb injury, with approximately $250,000$ instances occurring annually in the United States \cite{griffin2006}.   
Almost $100,000$ of these patients undergo ACL reconstruction (ACLR) surgery each year, which is accompanied by a rehabilitation program lasting several months \cite{wilk2012}.
While injuries resulting from contact are often beyond an individual's control, noncontact ACL ruptures occur most often when subjects exhibit biomechanical deficits during deceleration, lateral pivoting, and landing tasks \cite{besier2001, hewett2005, boden2010}.

These deficits are commonly diagnosed in functional motion screens \cite{padua2009, teyhen2014}, where a clinician looks for abnormal movement patterns and asymmetries in range of motion, strength, power, and control while the patient performs a series of dynamic tasks.
For ACLR patients, functional screens are used to track progress during rehabilitation and assess if an athlete is ready to return to sports \cite{manske2013, davies2017, gokeler2017}. 
However, since these assessments are not cost-effective for large groups or frequent sessions \cite{swart2014}, they cannot be performed regularly throughout the course of a rehabilitation program and are, in fact, inaccessible to most individuals. 


\begin{figure}[t]
    \centering
    \includegraphics[width = 0.5\textwidth]{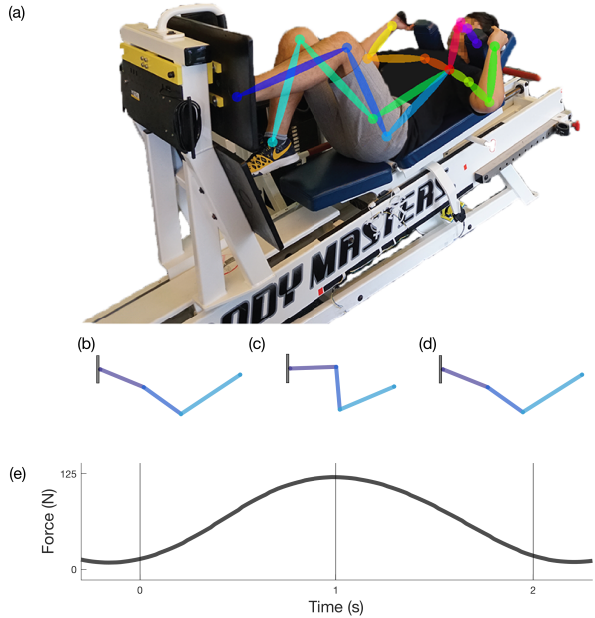}
    \caption{This paper employs a camera-based pose estimation algorithm to automatically estimate rehabilitation criteria for 12 ACL-reconstructed patients over the course of a 12-week study. (a) A stereo camera was used to collect video of participants performing a leg press exercise.
    A pose estimation algorithm estimated the positions of each of the participant's joints, depicted as a skeleton figure.
    (b, c, d) The estimated positions of the shank, thigh, and trunk are plotted at the beginning, middle, and end of a single leg press repetition, respectively.
    (e) The estimated force on the foot plate, which can be used to evaluate rate of recovery after an ACL reconstruction, (gray line) is plotted over time, with the vertical black lines corresponding to the times at which (b, c, and d) are plotted.}
    \label{fig:intro}
\end{figure}

Furthermore, functional screens, as well as some automated lower limb assessments \cite{mauntel2017}, focus on subjects' kinematics and typically do not evaluate forces. 
Measuring forces during motion could alert clinicians to altered load distributions throughout a patient's knee joint \cite{sanford2016}, but quantifying these forces is infeasible from clinical observations alone. 

To address these concerns, researchers have assessed high-risk movements quantitatively in the laboratory with motion capture equipment \cite{hewett2005, myer2014}.
Though successful in identifying biomechanical deficits that could lead to ACL rupture, the motion capture setup is inaccessible in the clinical setting due to cost and setup time.
Motion capture systems that involve equipment such as marker suits also inhibit a subject's natural movement \cite{atha1984, vlasic2007}, which can unintentionally impact the biomechanical assessment.
Additionally, automated progress assessment can be more easily implemented on a large scale by evaluating forces during commonly used rehabilitation exercises, such as the leg press \cite{neeter2006, esfandiarpour2013}.

The contributions of this paper are threefold. First, as described in Section \ref{sec:forceEst}, a technique to estimate forces through a vision-based system as subjects use a leg press in a rehabilitation program, outlined in Fig. \ref{fig:intro}.
Second, as described in Section \ref{sec:experiment}, a dataset that includes force, encoder, and camera data for $12$ ACLR subjects who were undergoing weekly rehabilitation for $12$ weeks.
Finally, as described in Section \ref{sec:results}, a quantitative evaluation that illustrates that our proposed vision-based technique is able to accurately measure common functional screening criteria when compared to empirical techniques and ground truth data across all participants.
The remainder of our paper describes quantifiable criteria present in the leg press functional screen (Section \ref{sec:background}) and discusses future work for automating other tests present in a functional motion screen (Section \ref{sec:conclusion}).


\section{Background} \label{sec:background}

This section describes criteria a clinician may look for when the leg press exercise is included in functional screenings for ACLR patients.
During functional motion screens, clinicians focus on a specific task with the intent of evaluating and tracking individual performance \cite{manske2013}.
In this paper, we focus on the single-leg leg press exercise, as it is widely used for rebuilding lower-limb muscle after surgery and can provide a means for assessing lower-limb symmetry \cite{neeter2006, esfandiarpour2013}.
During a leg press exercise, an individual lies on a sled that translates along a rail, and pushes on a foot plate to lift a set of adjustable weights.
While conducting a functional screen of an individual performing this task, clinicians can record the number of repetitions in each leg, comparing the patient's ACLR leg to the healthy one \cite{neeter2006, rohman2015}. 
Though research has shown that evaluating the force generated during the motion can be useful to assess ACL injury risk, clinicians are typically unable to accurately estimate this from direct observation \cite{sapega1990, thomee2011}.

Instruments such as a force plate and encoder can quantify the forces present in the leg press exercise as well as automate the recording of repetitions; however, these sensors are often expensive and require custom-built hardware to attach to a leg press.
Accordingly, this paper focuses on estimating the following quantities while an individual is performing a leg-press exercise, described in Section \ref{sec:forceEst}:
\begin{enumerate}
    \item Displacement of participant from initial position
    \item Force present on the leg press foot plate
    \item Symmetry in repetitions between left and right leg
\end{enumerate}
The above criteria are then used to track patient improvement over time.
We automate the estimation of these values using a single stereo camera coupled with a pose estimation algorithm, which does not require the use of expensive instrumentation or complicated installation.

\section{Automating the Estimation of Leg Press Functional Criteria} \label{sec:forceEst} 
This section details the method of data collection and processing for obtaining the quantities described in Section \ref{sec:background}. 
Here, familiarity with stereo camera geometry is assumed; for definitions refer to \cite{szeliski2010}.
We estimate two-dimensional joint position in the left and right frames of a stereo image. 
We then use this joint position data to estimate displacement and force with a dynamic model of a subject on a leg press.
The force and displacement values are compared to those measured by a rotary encoder and force plate.

\subsection{Data Collection}

\begin{figure}[tb]
    \centering
    \includegraphics[width = \linewidth]{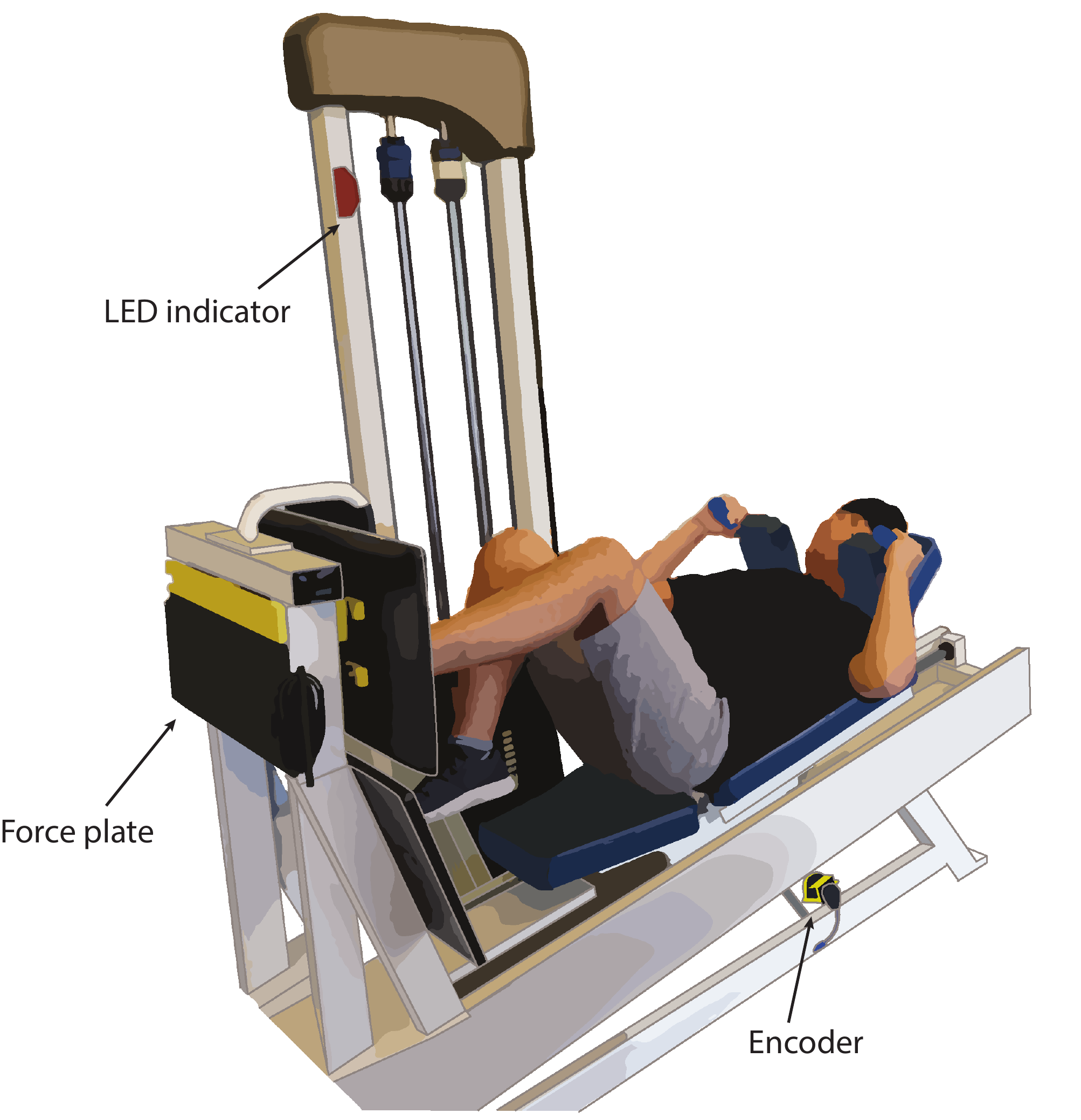}
    \caption{Setup for data collection, including instrumented leg press with force plate, encoder, and LED indicator to signal when a patient is at $90$ degrees. Video was recorded using a wall-mounted stereo camera (not pictured).}
    \label{fig:datacollection}
\end{figure}

Each test is performed on a leg press machine (Body Masters MD-122) instrumented with a force plate and encoder at University of Michigan's MedSport Facility in Ann Arbor, Michigan.
The force plate (Loadstar Sensors, DI-1000) with a sampling rate of $192$ Hz, is installed on the machine's foot plate, and measures the force exerted on the foot plate during the leg press test.
The rotary encoder (US Digital H6-10000), with sampling rate of $55$ Hz, is installed via an adapter on a pulley that rotates with the motion of the sled.
To record lateral movement of the subject, a stereo camera (ZED Stereolabs) collecting video at an average frequency of $8$ Hz is mounted on a wall near the leg press $4$ meters away. 
Mounting the camera on the wall ensures that the camera's position will not be disturbed by visiting patients over the course of the study. 
A magnetic switch is manually placed at the beginning of each test, triggering an LED light to ensure that subjects consistently reach a $90$-degree knee angle.
The experimental setup is shown in Fig. \ref{fig:datacollection}. 
Individual tests are eliminated if the camera, force plate, or encoder data fails to be recorded.

\subsection{Estimating Joint Position}


The OpenPose pose estimation algorithm is used to estimate the position of the hip joint during each leg press test \cite{cao2017}. 
This algorithm returns horizontal ($x$) and vertical ($y$) pixel coordinates for a joint of interest in each image frame.
It is run separately on rectified \cite[Chapter~11]{szeliski2010} left and right frames of the stereo images, and the $(x,y)$ coordinate pairs are saved for both frames. 
Although rectification guarantees that corresponding points in left and right images have the same $y$-coordinates, this property does not hold in the context of a pose estimation algorithm.
However, equivalent $y$-coordinates are necessary for accurate depth estimation on rectified images, due to the epipolar constraint \cite[Chapter~11]{szeliski2010}.
As the $y$-values of joint positions produced by OpenPose frequently differ between the left and right frame of the stereo images, we regulate the $y$-coordinates of corresponding left and right frames by replacing both values with their mean.

\subsection{Estimating Displacement}
\begin{figure}[tb]
    \centering
    \includegraphics[width = \linewidth]{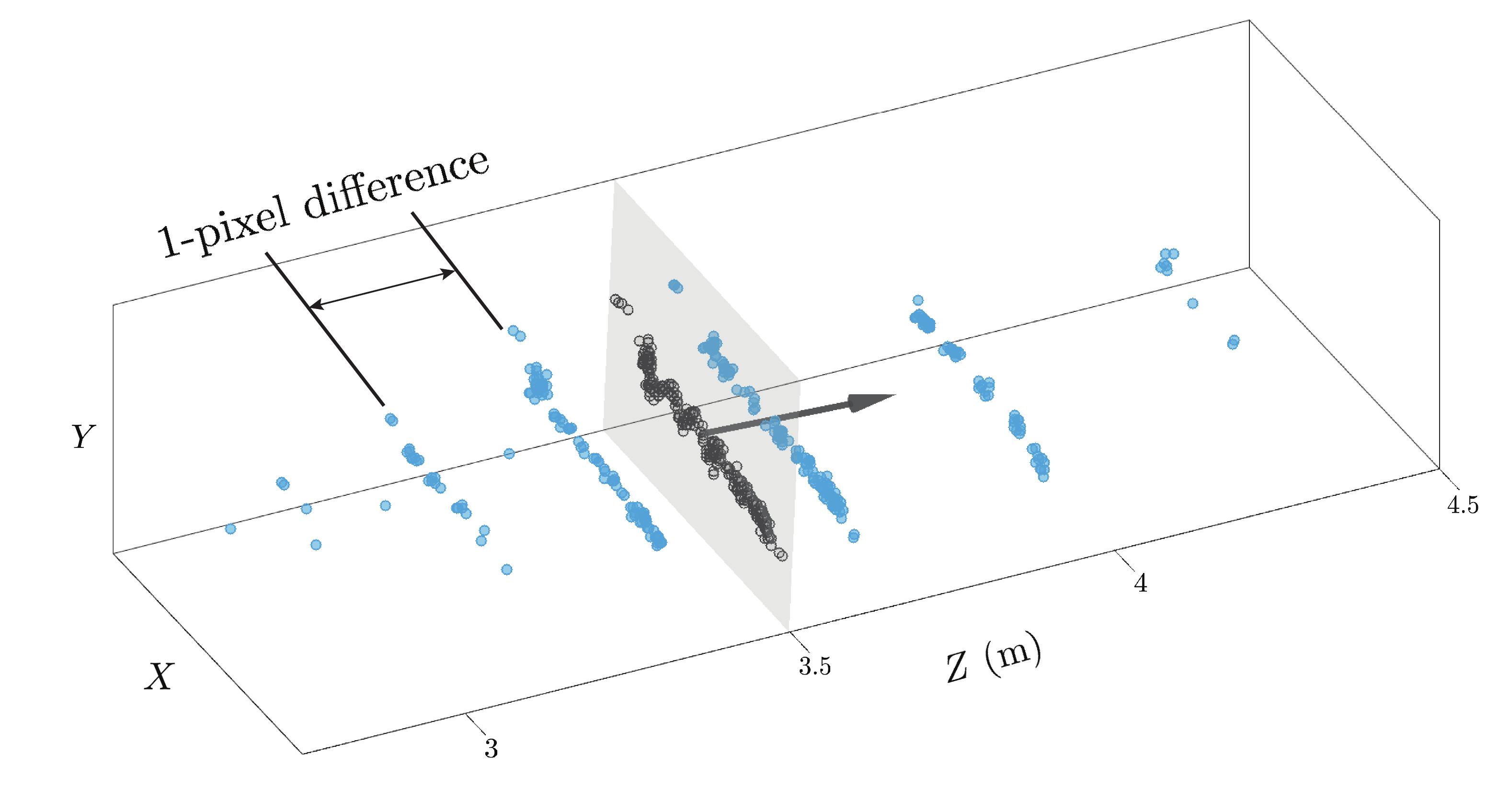}
    \caption{The estimated three-dimensional coordinates of the hip joint are plotted as blue dots in a camera-centric world frame $(X,Y,Z)$, all scaled equally. A Principal Component Analysis yields the first principal component of the data, shown as the gray arrow. The plane perpendicular to this vector is shown in light gray. The three-dimensional hip joint positions are projected onto this plane (gray dots), and then used to estimate a patient's displacement from their starting position.}
    \label{fig:PCA}
\end{figure}

To estimate a patient's displacement from their starting position using camera data, the three-dimensional position of each joint with respect to the left frame of the stereo camera is recovered. Using the post-processed $x$- and $y$-coordinates from our two-dimensional joint position, we compute a Direct Linear Transformation (DLT) with least squares minimization \cite[Chapter~7]{szeliski2010} and form a three-dimensional point cloud. An example from one patient can be seen in Fig. \ref{fig:PCA}.

As described in the previous section, regulation of a joint's $y$-coordinate lowers the re-
projection error \cite[Chapter~6]{szeliski2010} within corresponding stereo image pairs sampled at the same time.
However, the pose estimation algorithm may not return the same pixel for a given joint in image pairs sampled consecutively, meaning that even after regulation, the estimation of the depth of the joint relative to the camera can vary between successive frames.
Indeed, due to the distance of the camera from the study participants, error of even one pixel translates to a much larger error in the three-dimensional position estimation of the depth, or $Z$-coordinate, as illustrated in Fig. \ref{fig:PCA}.
The direction of this error is extracted via Principal Component Analysis (PCA) of the three-dimensional point cloud for the hip joint, where it is represented by the first principal component.
The second and third principal component then form a plane approximately containing the true motion of the hip joint, shown in Fig. \ref{fig:PCA}.
Exploiting the knowledge that a participant's hip joint position is constrained while performing a leg press and exhibits only minimal lateral movement, we project the three-dimensional points onto this plane.
The Euclidean distance between the resulting points and the starting position is then used to calculate displacement.



\subsection{Estimating Force}

The estimate of displacement is used to approximate the forces present on the leg press plate during testing. 
A dynamic model of the leg press machine is constructed to estimate the force on the foot plate at time $t$ by the patient, which we denote by $f(t)$.
The parameters employed by the model are:
\begin{conditions}
m     &  mass of patient\\
m_s     &  mass of sled \\
m_w     &  mass of weights\\
I   &   combined rotational inertia of pulleys\\
r_1     &   radius of pulley 1\\
r_2     &   radius of pulley 2\\
\alpha  &   angle of foot plate to vertical\\
\beta     &   angle of machine rail to horizontal\\
\theta &  rotation angle of the disk (encoder)\\
g   &   gravitational constant\\
\end{conditions}
Given the second derivative of a camera-estimated displacement trajectory $\ddot{x}(t)$, a free body diagram of the leg press machine, shown in Fig. \ref{fig:FBD}, yields: 
\begin{figure}[t]
    \centering
    \includegraphics[width = 0.45\textwidth]{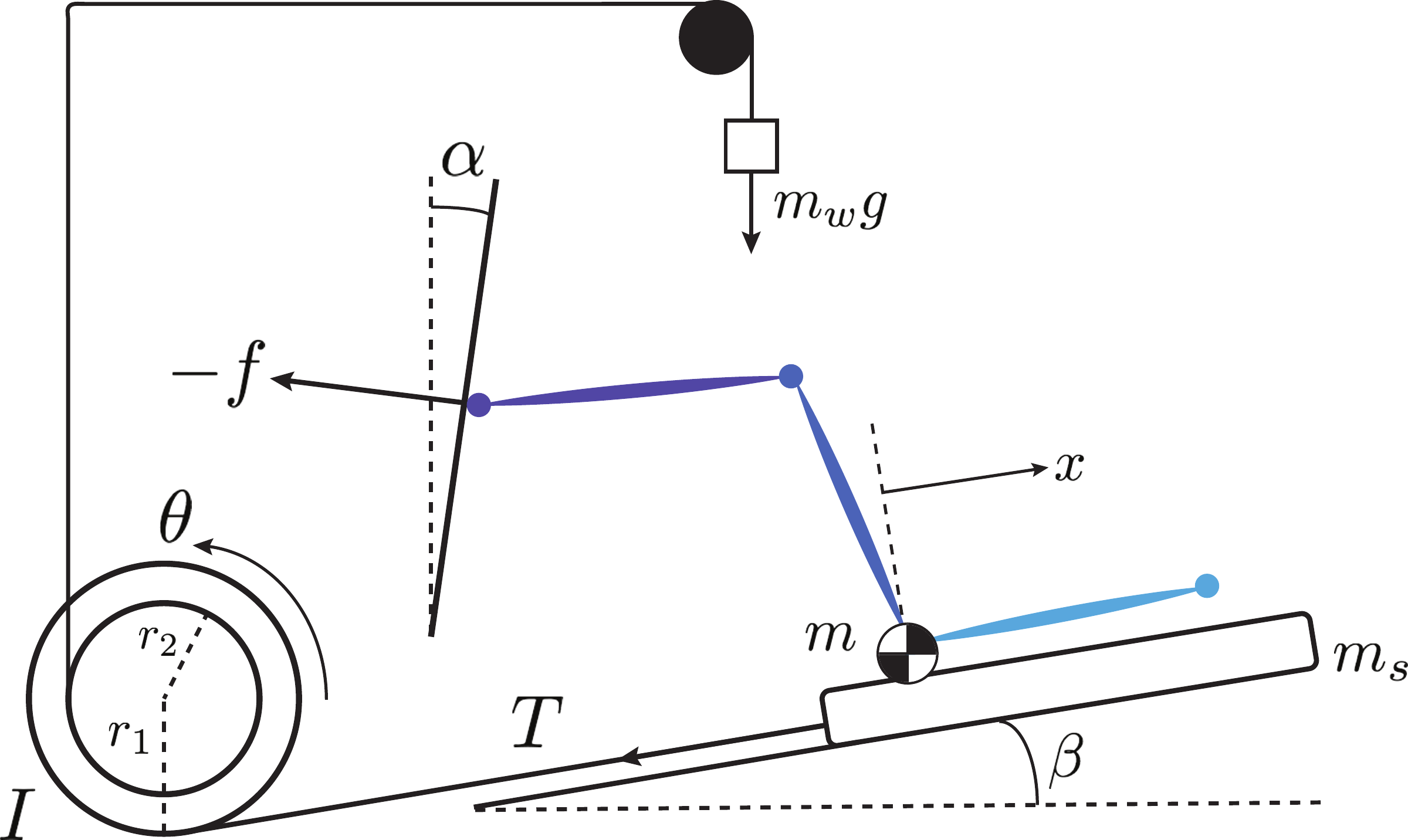}
    \caption{Free body diagram of the entire leg press system.}
    \label{fig:FBD}
\end{figure}
\begin{equation}
\label{eqn:forceplate-1}
    (m+m_s)\ddot{x}(t) = f(t)\cos(\alpha+\beta) - T - (m+m_s)g\sin{\beta}
\end{equation}
where $T$ represents the tension in a strap connecting the sled to the adjustable weights. 
This tension $T$ can be solved for through the following relation, where $\theta = \frac{x(t)}{r_1}$:

\begin{equation}
\label{eqn:forceplate-2}
  I\ddot{\theta}(t) = Tr_1-m_wgr_2
\end{equation}
This can be substituted into (\ref{eqn:forceplate-1}) to give the force $f(t)$:
\begin{equation}
\label{eqn:forceplate}
\begin{split}
    f(t) = \frac{1}{\cos(\alpha+\beta)}\bigg( \left((m + m_s) + \frac{I}{r_1^2} \right) \ddot{x}(t) + \\ + \frac{r_2}{r_1}m_w g + (m + m_s) g \sin{\beta}\bigg)
\end{split}
\end{equation}

\subsection{Calculating Repetitions and Symmetry}
The number of repetitions in each trial is calculated automatically by counting the peaks present in the smoothed estimation of displacement.
Within the time window of interest, the displacement data is first zero-centered by subtracting the mean value.
Next, the number of zero-crossings with negative slope are tallied as repetitions.
Defining the number of repetitions on the right and left leg, respectively, as $\text{reps}_R$ and $\text{reps}_L$, we calculate the symmetry between a patient's right and left legs as
\begin{equation}
    \label{eqn:percentsymmetry}
    \%_{sym} = 100\frac{\min{(\text{reps}_R, \text{reps}_L)}}{\max{(\text{reps}_R, \text{reps}_L)}}
\end{equation}

\section{Experiment} \label{sec:experiment}
This section describes the leg press experiment, conducted on the setup described in Section \ref{sec:forceEst}.
Twelve ACLR patients ($3$ female and $9$ male, ages $17$-$31$, average height $1.755 \pm 0.0876$ m, average body mass $78.14 \pm 10.7$ kg) were recruited for this study. 
The subjects were at least 4 months out of surgery. 
The same number of control subjects were recruited, matched with ACLR subjects for height, weight, sex, and age.
All participants gave their informed written consent, with parent or legal guardian permission if necessary.
The experimental protocol was approved by the University of Michigan Health Sciences and Behavioral Sciences
Institutional Review Board, eResearch ID: HUM00132288.
The ACLR subjects completed the leg press test once a week over a $12$-week period, and control subjects completed the test once.
During each testing session, the subject performed four $35$-second leg press intervals (single-leg, two intervals on each).
These intervals consisted of $30\%$ and $50\%$ of the subject's body weight on each leg, in randomized order. 
Subjects were instructed to complete as many repetitions of the leg press motion as possible, reaching a $90$-degree knee angle on each cycle.

\section{Results} \label{sec:results}
This section describes the accuracy of our method in estimating the following quantities: 
\begin{enumerate}
    \item Displacement of participant from initial position
    \item Force present on the leg press foot plate
    \item Symmetry in repetitions between left and right leg
\end{enumerate}
All accuracy measures were computed using force and displacement values from $505$ leg press tests, the total of $12$ ACLR patients over $12$ weeks with $4$ tests per session, as well as the $12$ controls patients with one session each.
As described earlier, any trials wherein the camera, force plate, or encoder data failed to record were excluded from analysis.
The estimated quantities are then used to track patient progress over time.

\subsection{Displacement Estimation}
\begin{figure}[ht]
    \centering
    \includegraphics[width = \linewidth]{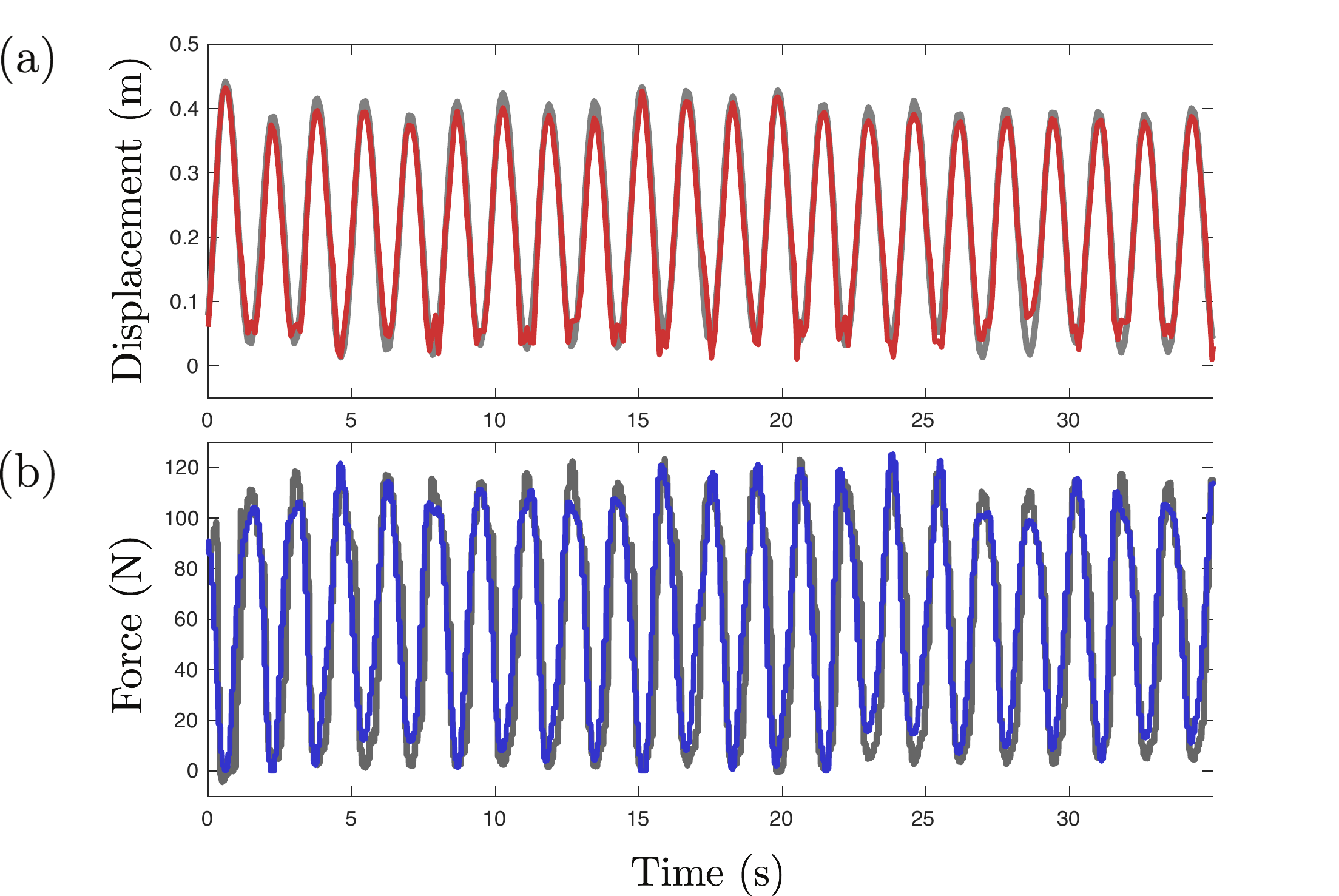}
    \caption{(a) Leg press displacement estimation (red line) versus measured (gray line) during the leg press test. (b) Leg press force plate value estimation (blue line) versus measured (gray line). } 
    \label{fig:estimation}
\end{figure}

The root mean square error (RMSE) between the estimated and measured displacement trajectory for each leg press trial was computed.
On average, the estimated displacement differed from measurements with an RMSE of $0.0411 \pm 0.0595$ m. 
When normalized by the range of measured displacements (mean $0.399$ m) in each trial, this corresponds to an average normalized root mean square error (NRMSE) of $10.3\% \pm 14.9\%$.
An example of estimated versus measured displacement values for one subject is shown in Fig. \ref{fig:estimation}, while Tab. \ref{results_table} reports a summary of estimation accuracy.

\begin{table}[t]
\caption{Accuracy of camera-based estimation}
\label{results_table}
\begin{center}
\begin{tabular}{|c||c|}
\hline
Number of Tests & 505 \\
\hline
Distance Estimation Accuracy(\%) & 89.7 \\
\hline
Force Estimation Accuracy (\%) & 85.3 \\
\hline
Reps Accuracy(\%) & 97.4 \\
\hline
Percent Symmetry Accuracy(\%) &  96.5 \\
\hline
\end{tabular}
\end{center}
\end{table}

\subsection{Force Estimation}
The RMSE between the estimated and measured force values for each leg press trial was found to be $14.48 \pm 4.93$ N.
After normalization by the range of measured forces, this corresponds to an NRMSE of $14.7\% \pm 4.5\%$.
An example of estimated versus measured force values for one subject is shown in Fig. \ref{fig:estimation}.
\subsection{Repetition Counting and Symmetry Evaluation}
A count of the number of repetitions completed in each of the $505$ trials was calculated using both estimated and measured displacement values.
The difference between the two counts was divided by the total number of repetitions found in the measured displacement trajectory, yielding an error percentage for each trial.
The average error percentage was $2.6\% \pm 1.7\%$.

The accuracy in percent symmetry was computed in a similar manner. After finding the repetitions from both the estimated and measured displacement values, percent symmetry was calculated for each using (\ref{eqn:percentsymmetry}). For each test, the difference between the estimated and measured percent symmetry values was divided by the measured percent symmetry. The error percentage, averaged across all tests, was $3.5\% \pm 2.3\%$.
\subsection{Patient Improvement Over Time}
\begin{figure}[b!]
    \centering
    \includegraphics[width = \linewidth]{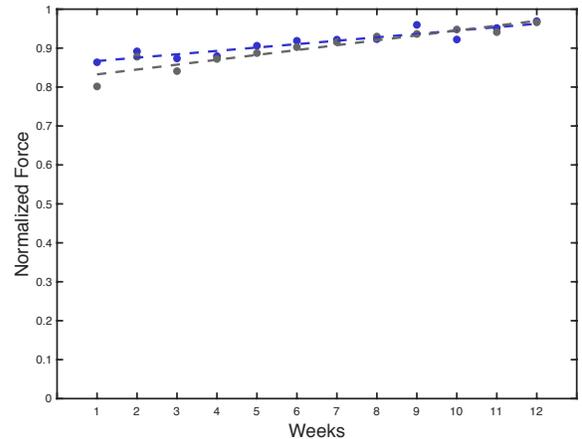}
    \caption{Normalized peak force, averaged across $12$ ACLR subjects, computed from both estimated (blue dots) and measured (gray dots) force values, plotted over a $12$-week period. Trendlines for the estimated and measured values are shown in the blue ($r^2 = 0.857$) and gray ($r^2 = 0.887$) dotted lines, respectively. The estimated average peak force increased by $9.5\%$ between the beginning and end of the study, while the measured data give an average increase of $13.7\%$.} 
    \label{fig:force_result}
\end{figure}

Force and displacement data collected from subjects' ACLR leg during the $50\%$ body weight test was used to track progress over time.
\begin{figure}[b!]
    \centering
    \includegraphics[width = \linewidth]{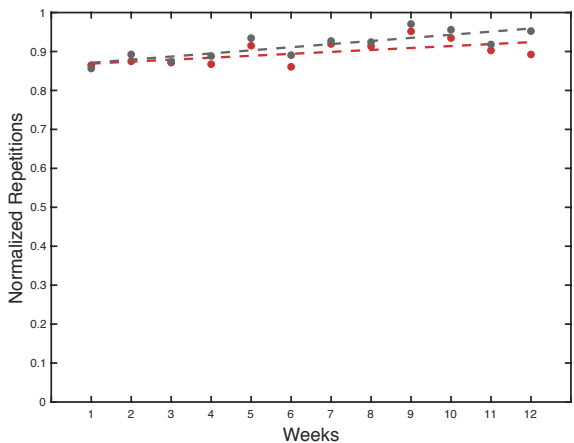}
    \caption{ Normalized number of repetitions averaged over $12$ ACLR patients, calculated with estimated (red dots) and measured (gray dots) displacement values, plotted over $12$ weeks. 
    Trendlines for the estimated and measured values are shown in red ($r^2 = 0.382$) and gray ($r^2 = 0.662$), respectively. The estimated data show an average increase in repetitions of $5.5\%$ over the course of the study, while the measured data give an average increase of $9.6\%$.}
    \label{fig:reps_result}
\end{figure}
The camera-based estimations of displacement and force confirm improvement in ACLR patient performance over the $12$-week study.
The number of repetitions achieved by each subject during each of the $12$ tests was normalized by the maximum number of repetitions for that subject over the course of the study.
Peak force was normalized in a similar fashion.
The $12$ ACLR patients in this program had an average increase in normalized number of repetitions of $5.5\%$, shown in Fig. \ref{fig:reps_result}, and average increase in peak force of $9.5\%$, illustrated in Fig. \ref{fig:force_result}.


\section{Discussion} \label{sec:conclusion}

Functional motion screens are useful for preventing ACL injury and monitoring rehabilitation progress after ACL reconstruction surgery. 
However, since they are time- and cost-intensive, these assessments have limited availability. 
Many of the qualitative criteria in a functional screen can be quantified, and introducing markerless pose estimation provides an opportunity to automate the process.

In this paper, we reproduced force and encoder readings from an instrumented leg press using a stereo camera with over $85\%$ accuracy.
We used these values to automate functional screening results. 
We calculate percent symmetry between a patient's healthy and ACL-reconstructed legs, with an accuracy over $95\%$.
We also tracked number of repetitions and peak force over the course of the $12$-week study for $12$ subjects.
The improvement in peak force suggests that measuring forces during ACLR rehabilitation may be clinically relevant for assessing if a patient is ready to return to sports.
However, more conclusive analyses of improvement trends in ACLR patients will be enabled through data currently being collected from a larger number of test participants.


These same estimated force and displacement values can contribute to future work in quantifying power and control during the leg press exercise.
Power is necessary for return to sports in ACLR patients \cite{thomee2011}, but is often qualitatively rated during functional screens.
Our force and displacement estimates can be used to quantify a patient's power by looking at force data over time.
To evaluate control, clinicians check whether patients are both achieving full extension and maintaining quadriceps activation without slamming weights, taking breaks, or hyper-extending the knee.
Our estimated displacement measurements can be used to detect breaks and knee hyper-extension during the leg press exercise.
Furthermore, the kinematic data recovered through the pose estimation algorithm for the $12$ ACLR subjects can be used in future work to quantify each subject's knee angles and estimate joint loads, as excessive valgus and varus loading in the knee joint are an indicator of ACL injury risk \cite{besier2001, hewett2005, boden2010}.


The results from this study highlight the effectiveness of a markerless, inexpensive camera-based method for tracking rehabilitation progress in ACLR patients, which can be extended to more functional tasks in the future. 
The leg press environment in this study allowed for simple dynamic modeling by constraining the axis of motion.
Similarly, progress assessment in other functional screening tasks, such as jump tests, lunges, or squats, can be automated by exploiting the structure in each exercise. 

\addtolength{\textheight}{-12cm}   

\section*{Acknowledgments}

The authors thank the test subjects and clinicians involved in the leg press tests.


\bibliographystyle{ieeetr}
\bibliography{references}

\begin{thebibliography}{10}

\bibitem{griffin2006}
L.~Griffin, M.~Albohm, and E.~Arendt, ``Understanding and preventing noncontact
  anterior cruciate ligament injuries: A review of the hunt valley ii
  meeting,'' {\em American Journal of Sports Medicine}, vol.~2006,
  pp.~1512--1532, 01 2006.

\bibitem{wilk2012}
K.~E. Wilk, L.~C. Macrina, E.~L. Cain, J.~R. Dugas, and J.~R. Andrews, ``Recent
  advances in the rehabilitation of anterior cruciate ligament injuries,'' {\em
  Journal of Orthopaedic \& Sports Physical Therapy}, vol.~42, no.~3,
  pp.~153--171, 2012.

\bibitem{besier2001}
T.~Besier, D.~Lloyd, J.~Cochrane~Wilkie, and T.~Ackland, ``External loading of
  the knee joint during running and cutting maneuvers,'' {\em Medicine and
  science in sports and exercise}, vol.~33, pp.~1168--75, 08 2001.

\bibitem{hewett2005}
T.~E. Hewett, G.~D. Myer, K.~R. Ford, J.~Robert S.~Heidt, A.~J. Colosimo, S.~G.
  McLean, A.~J. van~den Bogert, M.~V. Paterno, and P.~Succop, ``Biomechanical
  measures of neuromuscular control and valgus loading of the knee predict
  anterior cruciate ligament injury risk in female athletes: A prospective
  study,'' {\em American Journal of Sports Medicine}, vol.~33, no.~4,
  pp.~492--501, 2005.
\newblock PMID: 15722287.

\bibitem{boden2010}
B.~P~Boden, F.~Sheehan, J.~S~Torg, and T.~Hewett, ``Noncontact anterior
  cruciate ligament injuries: Mechanisms and risk factors,'' {\em The Journal
  of the American Academy of Orthopaedic Surgeons}, vol.~18, pp.~520--7, 09
  2010.

\bibitem{padua2009}
D.~Padua, S.~W~Marshall, M.~Boling, C.~Thigpen, W.~Garrett, and A.~I~Beutler,
  ``The {L}anding {E}rror {S}coring {S}ystem ({LESS}) is a valid and reliable
  clinical assessment tool of jump-landing biomechanics: The {JUMP}-{ACL}
  study,'' {\em American Journal of Sports Medicine}, vol.~37, pp.~1996--2002,
  10 2009.

\bibitem{teyhen2014}
D.~Teyhen, M.~Bergeron, P.~Deuster, N.~Baumgartner, A.~I~Beutler, S.~de~la
  Motte, B.~H~Jones, P.~Lisman, D.~Padua, T.~Pendergrass, S.~W~Pyne,
  E.~Schoomaker, T.~C~Sell, and F.~O'Connor, ``Consortium for health and
  military performance and american college of sports medicine summit: Utility
  of functional movement assessment in identifying musculoskeletal injury
  risk,'' {\em Current sports medicine reports}, pp.~52--63, 01 2014.

\bibitem{manske2013}
R.~Manske and M.~Reiman, ``Functional performance testing for power and return
  to sports,'' {\em Sports Health}, vol.~5, no.~3, pp.~244--250, 2013.
\newblock PMID: 24427396.

\bibitem{davies2017}
G.~J. Davies, E.~McCarty, M.~Provencher, and R.~C. Manske, ``{A}{C}{L} return
  to sport guidelines and criteria,'' {\em Current Reviews in Musculoskeletal
  Medicine}, vol.~10, pp.~307--314, Sep 2017.

\bibitem{gokeler2017}
A.~Gokeler, W.~Welling, S.~Zaffagnini, R.~Seil, and D.~Padua, ``Development of
  a test battery to enhance safe return to sports after anterior cruciate
  ligament reconstruction,'' {\em Knee Surgery, Sports Traumatology,
  Arthroscopy}, vol.~25, pp.~192--199, Jan 2017.

\bibitem{swart2014}
E.~Swart, L.~Redler, P.~D~Fabricant, B.~R~Mandelbaum, C.~Ahmad, and
  Y.~Claire~Wang, ``Prevention and screening programs for anterior cruciate
  ligament injuries in young athletes: A cost-effectiveness analysis,'' {\em
  American Journal of Bone and Joint Surgery}, vol.~96, pp.~705--11, 05 2014.

\bibitem{mauntel2017}
P.~A. Mauntel, Timothy~C, P.~A. Padua, Darin~A, D.~P.~S. Stanley, Laura~E,
  P.~A. Frank, Barnett~S, P.~A. DiStefano, Lindsay~J, M.~A.~C. Peck, Karen~Y,
  P.~M.~A. Cameron, Kenneth~L, and P.~Marshall, Stephen~W., ``Automated
  quantification of the landing error scoring system with a markerless
  motion-capture system,'' {\em Journal of Athletic Training}, vol.~52,
  pp.~1002--1009, 11 2017.

\bibitem{sanford2016}
B.~A. Sanford, J.~L. Williams, A.~Zucker-Levin, and W.~M. Mihalko, ``Asymmetric
  ground reaction forces and knee kinematics during squat after anterior
  cruciate ligament ({A}{C}{L}) reconstruction,'' {\em The Knee}, vol.~23,
  no.~5, pp.~820 -- 825, 2016.

\bibitem{myer2014}
G.~Myer, K.~Ford, S.~Di~Stasi, K.~Barber~Foss, L.~J~Micheli, and T.~Hewett,
  ``High knee abduction moments are common risk factors for patellofemoral pain
  ({P}{F}{P}) and anterior cruciate ligament ({A}{C}{L}) injury in girls: Is
  {P}{F}{P} itself a predictor for subsequent {A}{C}{L} injury?,'' {\em British
  Journal of Sports Medicine}, vol.~49, 03 2014.

\bibitem{atha1984}
J.~Atha, ``Current techniques for measuring motion,'' {\em Applied Ergonomics},
  vol.~15, no.~4, pp.~245 -- 257, 1984.

\bibitem{vlasic2007}
D.~Vlasic, R.~Adelsberger, G.~Vannucci, J.~Barnwell, M.~Gross, W.~Matusik, and
  J.~Popovi\'{c}, ``Practical motion capture in everyday surroundings,'' {\em
  ACM Trans. Graph.}, vol.~26, July 2007.

\bibitem{neeter2006}
C.~Neeter, A.~Gustavsson, P.~Thome{\'e}, J.~Augustsson, R.~Thome{\'e}, and
  J.~Karlsson, ``Development of a strength test battery for evaluating leg
  muscle power after anterior cruciate ligament injury and reconstruction,''
  {\em Knee Surgery, Sports Traumatology, Arthroscopy}, vol.~14, pp.~571--580,
  Jun 2006.

\bibitem{esfandiarpour2013}
F.~Esfandiarpour, A.~Shakourirad, S.~T. Moghaddam, G.~Olyaei, A.~Eslami, and
  F.~Farahmand, ``Comparison of kinematics of {A}{C}{L}-deficient and healthy
  knees during passive flexion and isometric leg press,'' {\em The Knee},
  vol.~20, no.~6, pp.~505 -- 510, 2013.

\bibitem{rohman2015}
E.~Rohman, J.~Steubs, and M.~Tompkins, ``Changes in involved and uninvolved
  limb function during rehabilitation after anterior cruciate ligament
  reconstruction,'' {\em American Journal of Sports Medicine}, vol.~43,
  pp.~1391--1398, 2015.

\bibitem{sapega1990}
A.~A. Sapega, ``Muscle performance evaluation in orthopaedic practice,'' {\em
  Journal of Bone and Joint Surgery}, vol.~72-A, pp.~1104--1110, 12 1990.

\bibitem{thomee2011}
R.~Thomee, Y.~Kaplan, J.~Kvist, G.~Myklebust, M.~Risberg, D.~Theisen,
  E.~Tsepis, S.~Werner, B.~Wondrasch, and E.~Witvrouw, ``Muscle strength and
  hop performance criteria prior to return to sports after {A}{C}{L}
  reconstruction,'' {\em Knee surgery, sports traumatology, arthroscopy :
  official journal of the ESSKA}, vol.~19, pp.~1798--805, 09 2011.

\bibitem{szeliski2010}
R.~Szeliski, {\em Computer Vision: Algorithms and Applications}.
\newblock London: Springer-Verlag, 2011.

\bibitem{cao2017}
Z.~Cao, T.~Simon, S.-E. Wei, and Y.~Sheikh, ``Realtime multi-person 2{D} pose
  estimation using part affinity fields,'' in {\em CVPR}, 2017.

\end{thebibliography}

\end{document}